# A NOTE ON ACTIVE LEARNING FOR SMOOTH PROBLEMS

SATYAKI MAHALANABIS
SMAHALAN@CS.ROCHESTER.EDU

ABSTRACT. We show that the disagreement coefficient of certain smooth hypothesis classes is $O(m)$, where $m$ is the dimension of the hypothesis space, thereby answering a question posed in [Fri09].

We answer a question posed in [Fri09] regarding the disagreement coefficient of certain smooth hypothesis classes. To be precise, we show that the limiting disagreement coefficient, defined in Lemma 3 of [Fri09], for hypotheses and distributions as specified in Theorem 4 of [Fri09] is at most $2\sqrt{\frac{\pi}{2(1-\frac{1}{m})}}m \leq 2\sqrt{\pi}m$, where $m \geq 2$ is the dimension of the hypothesis space, thereby improving on the $2m^{3/2}$ bound given there. Our proof is exactly the same as Theorem 4 of [Fri09], except that we need to use the following proposition.

**Proposition 1.** *Consider a set of vectors $V \subseteq \mathbb{R}^m$ ($m \geq 2$), and let $K_m \subseteq \mathbb{R}^m$ be a symmetric, origin-centred and convex body. Then*

$$\frac{\sum_{v \in V} \sup_{h \in K_m} |v^t\, h|}{\sup_{h \in K_m} \sum_{v \in V} |v^t\, h|} \;\leq\; \sqrt{\frac{\pi}{2(1-\frac{1}{m})}}\; m. \tag{1}$$

The convex set $K_m$ and the vectors $V$ in Proposition 1 correspond respectively to the ball in $\mathbb{R}^m$ w.r.t the distance $\hat{d}$ and the vectors $\{a_x\}$ as defined in the proof of Theorem 4 in [Fri09]. Roughly speaking, $K_m$ is the version space. Note that the limiting disagreement coefficient is actually twice the l.h.s of (1).

**Proof :**
(Proposition 1) As in [Fri09], consider the John ellipsoid $\mathcal{E} = \{x^t A^t A x \leq 1\}$ s.t. $\mathcal{E} \subseteq K_m \subseteq \sqrt{m}\mathcal{E}$. Then clearly we have

$$\frac{\sum_{v \in V} \sup_{h \in K_m} |v^t\, h|}{\sup_{h \in K_m} \sum_{v \in V} |v^t\, h|} \;\leq\; \sqrt{m}\frac{\sum_{v \in V} \sup_{h \in \mathcal{E}} |v^t\, h|}{\sup_{h \in \mathcal{E}} \sum_{v \in V} |v^t\, h|} \;=\; \sqrt{m}\frac{\sum_{v \in V} \sup_{h \in S_{m-1}} |(v^t A^{-1})\, h|}{\sup_{h \in S_{m-1}} \sum_{v \in V} |(v^t A^{-1})\, h|}, \tag{2}$$

where $S_{m-1}$ is the (surface of) $m$-dimensional unit sphere. Now if we choose $h$ from the uniform distribution $U_m$ on $S_{m-1}$, we have

$$\sup_{h \in S_{m-1}} \sum_{v \in V} |(v^t A^{-1})\, h| \;\geq\; \mathop{\mathrm{E}}_{h \sim U_m}\left[\sum_{v \in V} |(v^t A^{-1})\, h|\right] \;=\; \sum_{v \in V} \mathop{\mathrm{E}}_{h \sim U_m}\left[|(v^t A^{-1})\, h|\right] \;\geq\; \sum_{v \in V} \frac{c_m}{\sqrt{m}} \|v^t A^{-1}\|, \tag{3}$$

where $c_m = \sqrt{\frac{2}{\pi}(1-\frac{1}{m})}$ and where we have used the fact (see e.g. [Bau90]) that for any unit vector $u$, $\mathop{\mathrm{E}}_{h \sim U_m}\left[|u^t\, h|\right] \geq c_m / \sqrt{m}$. Note also that for each $v$, $\sup_{h \in S_{m-1}} |(v^t A^{-1})\, h| = \|v^t A^{-1}\|$. Hence (1) follows by substituting (3) in (2). ∎

**Remark 2.** The l.h.s of (1) is $m$ for the example of the origin-centred $m$-dimensional octagon (i.e the convex hull of $2m$ points $\{[\pm 1 \ 0 \ \ldots \ 0]^t, \ [0 \ \pm 1 \ \ldots \ 0]^t, \ldots, \ [0 \ 0 \ \ldots \ \pm 1]^t\}$) and the set of $m$ vectors $\{[1 \ 0 \ \ldots \ 0]^t, \ [0 \ 1 \ \ldots \ 0]^t, \ldots, \ [0 \ 0 \ \ldots \ 1]^t\}$. Tightening the bound in (1) to $m$ would probably need more careful analysis.